\documentclass[10pt,twocolumn,letterpaper]{article}

\usepackage{wacv}
\usepackage{times}
\usepackage{epsfig}
\usepackage{graphicx}
\usepackage{amsmath}
\usepackage{amssymb}
\usepackage[english,algoruled,linesnumbered,vlined]{algorithm2e}
\let\oldnl\nl% Store \nl in \oldnl
\newcommand{\nonl}{\renewcommand{\nl}{\let\nl\oldnl}}% Remove line number for one line

\usepackage{url}
\usepackage[hyperfootnotes=false]{hyperref}

\usepackage{wrapfig}
\usepackage{float}
\usepackage{tabu,multirow}
\restylefloat{table}
\usepackage{placeins}
\usepackage{booktabs}
\usepackage{footnote}
\usepackage{tablefootnote}
\usepackage{afterpage}

\usepackage[table,xcdraw]{xcolor}

\usepackage{mathtools}  
\mathtoolsset{showonlyrefs}

\usepackage[caption=false]{subfig} % for subfigures
\usepackage{siunitx} % used for degrees symbol

\newlength{\colwidth}
\newcommand{\todo}[1]{}
\renewcommand{\todo}[1]{{\color{red} TODO: {#1}}}

\usepackage{manyfoot}

\DeclareNewFootnote{A}
\DeclareNewFootnote{B}

\let\footnoteR\footnoteB
\let\footnote\footnoteA

\wacvfinalcopy % *** Uncomment this line for the final submission

\def\wacvPaperID{519} % *** Enter the wacv Paper ID here

\ifwacvfinal\fi
\hypersetup{draft}
\begin{document}

%%%%%%%%% TITLE
\title{Human-centric light sensing and estimation from RGBD images: \\ The invisible light switch} %Is 3D modeling of a real scene with a single \\camera sufficient for reliable light modeling? TOCHANGE}

% Authors at the same institution
%\author{First Author \hspace{2cm} Second Author \\
% Authors at different institutions
% \author{First Author \\
% \and
% Second Author \\

\author{Theodore Tsesmelis$^{1,2,3}$, Irtiza Hasan$^{3,1}$, Marco Cristani$^{2,3}$, Alessio Del Bue$^{2, \dagger}$, Fabio Galasso$^{1, \dagger}$\\
{\normalsize Corporate Innovation OSRAM GmbH$^{1}$, Istituto Italiano di Tecnologia (IIT)$^{2}$, University of Verona (UNIVR)$^{3}$}\\
{\tt\small t.tsesmelis@osram.com, irtiza.hasan@univr.it}}

\maketitle
\ifwacvfinal\thispagestyle{empty}\fi

\begin{abstract} \label{sec:abstract}
Lighting design in indoor environments is of primary importance for at least two reasons: 1) people should perceive an adequate light; 2) an effective lighting design means consistent energy saving.
We present the \emph{Invisible Light Switch} (ILS) to address both aspects. ILS dynamically adjusts the room illumination level to save energy while maintaining constant the light level perception of the users. So the energy saving is invisible to them.
Our proposed ILS leverages a radiosity model to estimate the light level which is perceived by a person within an indoor environment, taking into account the person position and her/his viewing frustum (head pose).
ILS may therefore dim those luminaires, which are not seen by the user, resulting in an effective energy saving, especially in large open offices (where light may otherwise be ON everywhere for a single person).
To quantify the system performance, we have collected a new dataset where people wear luxmeter devices while working in office rooms. The luxmeters measure the amount of light (in Lux) reaching the people gaze, which we consider a proxy to their illumination level perception. Our initial results are promising: in a room with 8 LED luminaires, the energy consumption in a day may be reduced from 18585 to 6206 watts with ILS (currently needing 1560 watts for operations). While doing so, the drop in perceived lighting decreases by just 200 lux, a value considered negligible when the original illumination level is above 1200 lux, as is normally the case in offices.

\end{abstract}

\section{Introduction}
\footnoteR{$^{\dagger}$These two authors contribute equally to the work.}
People generally do not consider the impact that indoor illumination has on the monthly costs of large environments such as offices or warehouses. At the same time, they are pretty sensible to illumination, especially during office activities such as drawing and studying or doing precision works. As a consequence, office illumination is often always ON, at the maximum available lighting level, which increases the energy consumption.      

\begin{figure}[!htb]
    % \vspace{-5pt}
	\begin{center}
	    \includegraphics[width=0.5\linewidth]{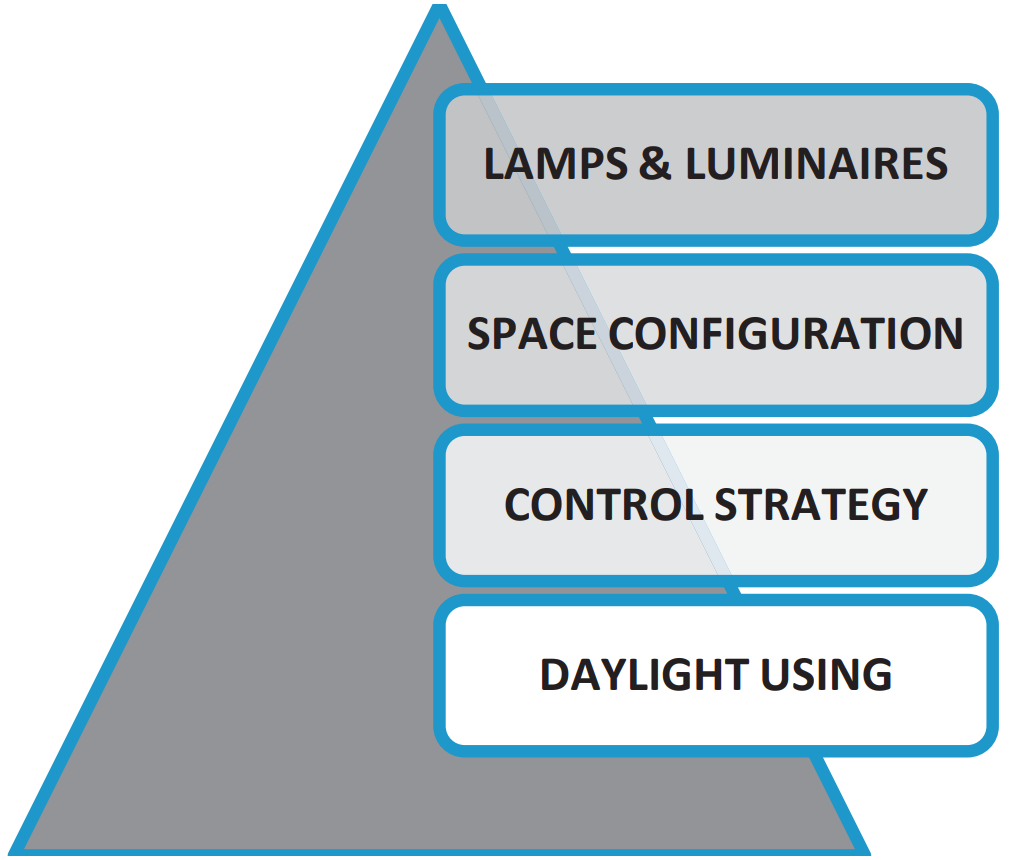}
	\end{center}
	\caption{The pyramidal scheme shows energy saving strategies \cite{tak2009}. From the bottom to the top this implies the exploitation of natural light, the distribution and control of lighting locally, efficient space configuration with bright colors and open spaces for better light propagation and utilization of the light sources driven by energy saving customization. The order of the strategies in the pyramid shows the importance that each action should be applied in a green-oriented building maintenance. }\label{fig:strategies}
 	\vspace{-2pt}
\end{figure}

The works of Kralikova and Zhou \etal \cite{kralikova2015energy,zhou2015data}, show that the lighting consumption of a building can take more than 15\% of the overall electricity consumption. While at peak periods, this can reach up to approximately a fourth or even more. It is clear that savings in the lighting are usually most evident and most easily feasible especially in environments where the human occupancy is limited. However, in dynamic environments where the human presence is more evident the power saving strategies are becoming more complex and harder to be addressed. The base energy saving techniques and strategies usually focus on the following principles: \textbf{a)} maximise the use of daylight; \textbf{b)} make lighting control as local as possible and get staff involved in energy saving planning; \textbf{c)} use bright coloured walls and ceilings; \textbf{d)} utilize and adjust the light sources to the most energy efficient lamp/luminaire combinations (see Figure \ref{fig:strategies}).

\begin{figure*}
    \begin{center}
        \includegraphics[width=0.85\linewidth]{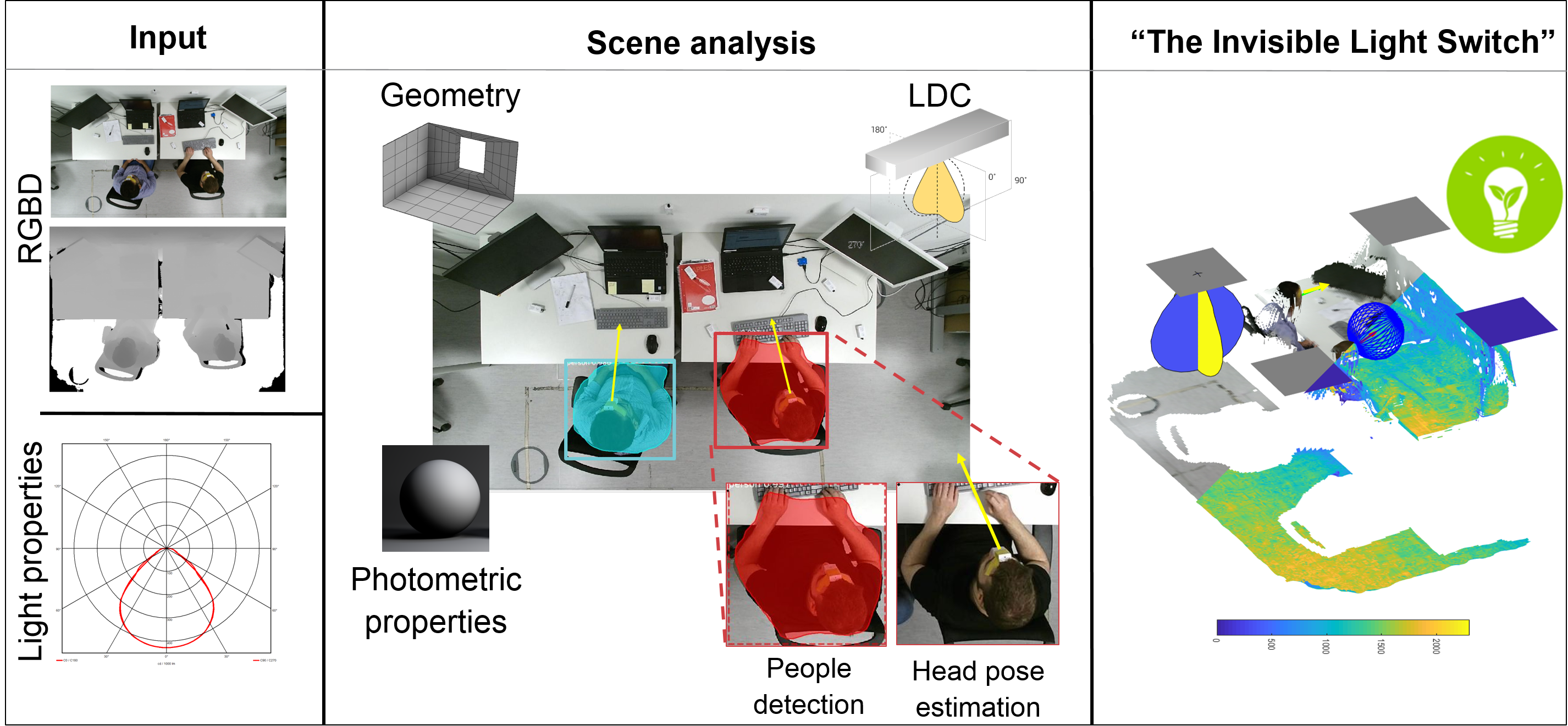}
    \end{center}
       \caption{Overall pipeline of our system. We first acquire the RGBD input from the camera system \textit{(left)} and together with the lighting system properties we use this information to create the Invisible Light Switch (ILS). That is, structuring the geometry of the scene, extracting the photometric properties of the material and applying a human centric analysis from where we detect the human presence in the scene and extract the possible head poses. Lastly we utilize the output of the scene analysis as the "Invisible Light Switch" application targeting a power saving framework.}
    \label{fig:teaser}
    \vspace{-5pt}
\end{figure*}

However, lighting can be used for much more than to illuminate. It can enhance productivity, creating flexible spaces that adapt to the task at hand. Energy-efficient lighting solutions for industry can reduce environmental impact and save on costs, while at the same time increasing the life quality and productivity.%%MARCO: references on this

\begin{figure}[!ht]
	\begin{center}
	    \includegraphics[width=0.69\linewidth]{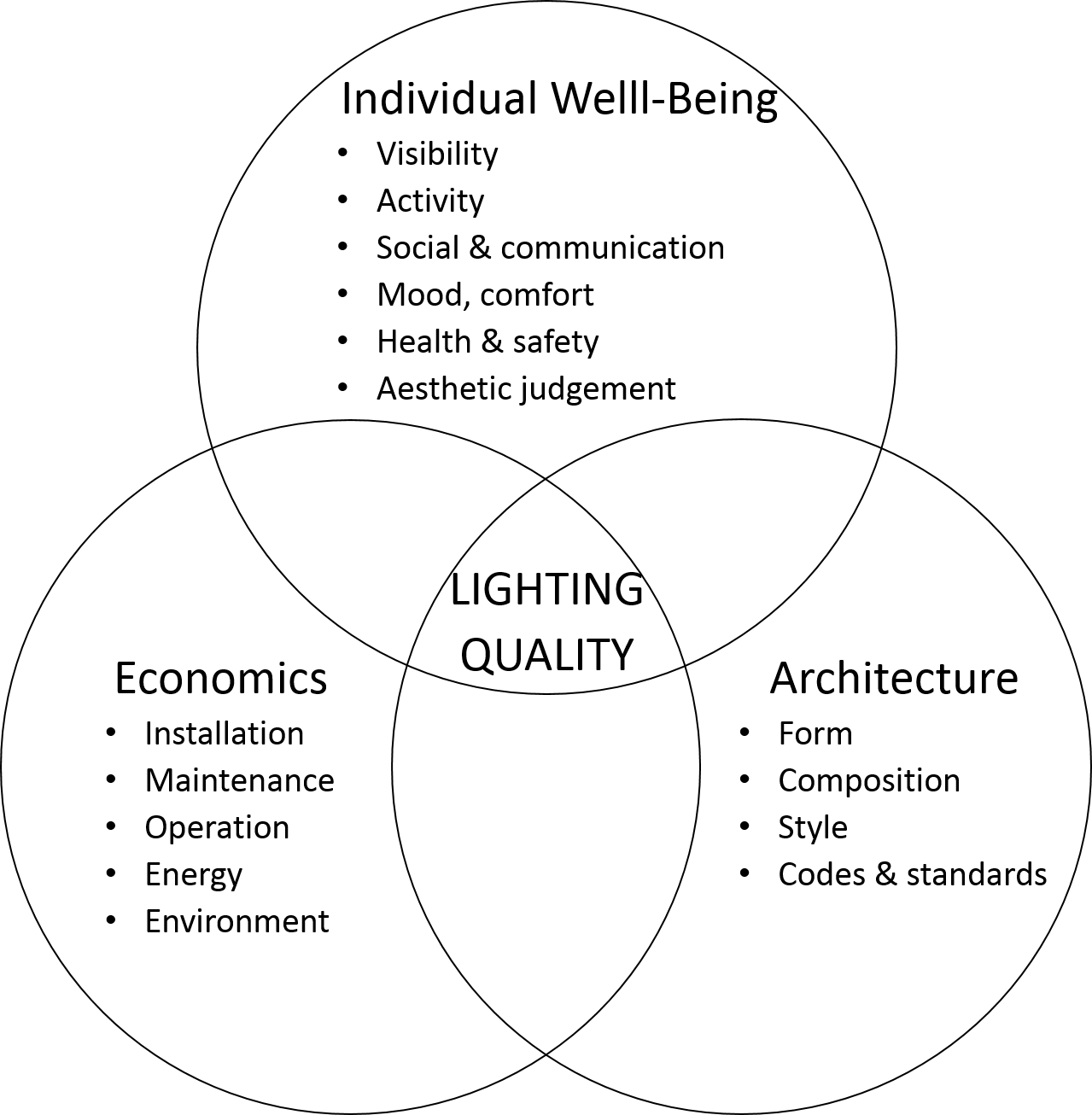}
	\end{center}
	\caption{Framework of lighting quality according to the International Association of Lighting Designers (IALD) \cite{iald2018}.}\label{fig:light_quality}
 	\vspace{-2pt}
\end{figure}

\iffalse 
%%MARCO: references on this
Good lighting in the workplace with well-lit task areas is essential for optimizing visual performance, visual comfort and ambience, especially with an aging workforce. 
%%MARCO: reference
How someone would define good lighting? light quality has been approached with different definitions. However, the definition that seems most generally applicable is that lighting quality is given by the extent to which the light sources installation meets the objectives and constraints set by the client and the designer. In this way, lighting quality is related to objectives like enhancing performance of relevant tasks, creating specific impressions, generating desired pattern of behaviour and ensuring visual comfort \cite{boyce2004lighting, gligor2006considerations}.
\fi
The International Association of Lighting Designers \cite{iald2018} states that optimal lighting consists of achieving an optimal balance among \textit{human needs, architectural considerations, and energy efficiency}  (Fig. \ref{fig:light_quality}).  

In this paper we present the Invisible Light Switch (ILS), a smart lighting framework for dynamically adjusting the illumination level in an indoor environment. ILS takes into account the geometry of the scene, the presence of people and their light perception with the goals of maximizing the human comfort in terms of perceived light and, at the same time,  with the lowest cost in terms of energy consumption.
Our framework builds upon a light estimation system capable of estimating the light in a given 3D point of a multi luminaire indoor environment \cite{tsesmelis2018}. In this work \cite{tsesmelis2018}, the radiosity model has been customized to take into account a realistic model of light propagation, outclassing even industrial software in the task.    

This paper enriches the model by including the human aspect, and showing how the interplay between the light estimation system and the human activity may lead to a consistent energy saving framework. The invisible light switch summarises the idea: an individual has the feeling of an environment which is globally illuminated, while in reality an automated light switch dims the luminaires in a way which is invisible to the users. This is possible by estimating the position of a person in the sensed environment, its head orientation, and understanding the light which is perceived by him. In fact, the lighting sensed by a human can be assumed as the light contained in a conic volume departing from the mean point connecting the human’s eyes in the direction of the nose. Given this, it is possible to determine which luminaries could be switched off/dimmed down while maintaining the level of perceived light unchanged. 
The head pose is provided by detecting the person first and then estimating the head orientation. The former is carried out by means the state-of-the-art detector Mask R-CNN \cite{he2017mask} with ResNet \cite{he2016deep} as a backbone architecture, while head pose is done using Hasan's \etal method \cite{hasan2017tiny}.

To test the system, a novel dataset has been built where 2 people are present in two rooms with 2 portable luxmeters attached to their forehead, well suited to mimic the human perception. This provides us with the ground truth information which can be considered when evaluating the light estimations of the system, given a particular setup of the luminaires. 
Experiments show how reliable is the system in detecting the people position along with head orientations and the related perceived light. A margin of 100 Lux error is observed within a global illumination estimation of 1200 lux. As reported by \cite{iesna2000}, this delta is barely perceivable by the human eye, so the error can be considered within the accepted range. Thereafter, it is shown that the ILS allows to heavily modify the illumination setup by only affecting the human perception of a delta of 200 Lux, still within the range of non-perceivable changes. We finally show that our system is promising in terms of energy saving: since in the most aggressive scenario could indicate for up to 66\% power efficiency.

The rest of the paper is organized as follows, next we review related work (Sec.~\ref{sec:relatedwork}) and define the overall proposed pipeline (Sec.~\ref{sec:method}). We present the results and evaluation in (Sec.~\ref{sec:evaluation}). Finally, we conclude in (Sec.~\ref{sec:conclusion}).

\section{Related Work}\label{sec:relatedwork}

\subsection{Lights and Behaviour}
Relationship between human activities and lights is a widely studied topic in perceptual sciences \cite{adams1991effect,flynn1979guide,gifford1988light}. Recently, it was shown by \cite{xu2013incandescent} that light intensifies people's perception. It triggers emotional system leading to intensified effective reactions. Light changes our perception of space \cite{galasiu2006occupant}, we tend to associate different illumination patterns to different social gatherings (musical concert \vs candle light dinner). People seem to share more details in bright light than darkness \cite{carr1974effects}, we as human beings also rely on facial expressions which are only visible in light. Moreover, light provides sense of security \cite{gifford1988light}, people choose roads and streets in night due to the illumination \cite{taylor1974movement}. Recently, studies targeting the office environments revealed a strong connection between people's productivity and the lights \cite{kralikova2015energy,kralikova2016lighting,smolders2012need}. Eyeing the importance of lighting on humans, related communities such as Human Computer Interaction (HCI) \cite{poulsen2013responsive} deployed interactive lighting in a city square, providing a sense of ``belongingness'' to the residents. Furthermore, ubiquitous computing \cite{gil2011experiences} and architectural design \cite{magielse2013designing} have also investigated this topic. However, there are also studies that question the relationship between the light perception and the actual measured spatial illumination \cite{bernecker2017,iesna2000}. 

\subsection{Modelling human activities}
Despite receiving a wide scale attention, the literature in computer vision seems to have ignored the modelling of light and behaviour. Only recently Hasan and Tsesmelis \etal \cite{hasan2017tsesmelis} presented the idea of jointly modeling the relationship of light and human behavior via long term time-lapse observation of the scene by recognizing and forecasting activities using the head pose estimation as a proxy for the gaze. 

% In this work, we propose the use of visual frustum of attention (VFOA) for scene understanding, activity recognition and activity forecasting. VFOA identifies the volume of a scene where fixation of a person may occur; It can be inferred from head pose estimation, and it is crucial in scenarios such as top-view office cameras and surveillance scenarios where precise gazing information cannot be retrieved.

Estimation of head pose is inherently a challenging task due to subtle differences between human poses. However, in the past several techniques ranging from low level image features to appearance based learning architectures were used to address the problem of head pose estimation. Previously, \cite{gourier2006head,voit2006bayesian} used neural networks to estimate head pose. While authors in \cite{benfold2009guiding} adopted a randomized fern based approach to estimate head orientation. Limited accuracy was achieved though due to several reasons such as two images of the same person in different poses appeared more similar than two different people in same pose or due to difficulty to compute low level image features in low resolution images. Recently, decision trees have been reported to achieve state of the art results \cite{lee2015fast}. However, they rely on local features and are prone to make errors when tested in real world crowded scenarios. We address the issue of having a head pose estimator that can work in unconstrained real world scenarios by utilizing the power of deep neural network models which in recent past, it has been used for pose estimation \cite{toshev2014deeppose}.

Having a strong similarity with head pose, some studies focused on estimating visual frustum (VFOA) on low resolution images \cite{ba2004probabilistic,RobertsonReidGazeLow2006,stiefelhagen1999gaze,stiefelhagen1999gaze} together with the general pose of the person. VFOA has been used as a reliable cue for identifying social interactions: in \cite{bazzani2013social} the head direction is used to estimate a 3D visual frustum as approximation of the VFOA of a person. Given the VFOA and spatial layout, human-human interactions are estimated: the concept is that people who are in a close proximity and having their VFOA intersecting with each other are engaged into a human-human interaction. The same concept has bee studied in \cite{RobertsonReid}. On the other hand, in \cite{smith2008tracking}, the VFOA was defined as a line directed towards the focus of attention by taking into account subject's gaze in low resolution images: in that work the goal was to understand the gazing behavior of people in front of a shop window. The VFOA was projected on the floor and modeled as a Gaussian distribution containing ``samples of attention'' in front of a person \cite{Cristani:FF:BMVC:2011}: the higher concentration, depicts stronger likelihood that in that area the eyes' fixation would be present. In a physiologically motivated study \cite{vascon2016cviu} the VFOA is represented by an angle $\theta$ (head orientation), an aperture $\alpha = 160 $ $degree$ and a length $l$. The $\theta$  corresponds to the variance of the Gaussian distribution modelled around the spatial proximity of a person. In the same study, the density of attention was used to measure the likelihood of a visual fixation: a more concentrated sampling was conducted at locations closer to the person, smoothly decreasing the frequency of sampling as one goes away. The visual frustum is constructed by sampling from the above Gaussian kernel and only considering ones confining inside the cone of attention composed by the angle $\alpha$. Finally in \cite{zhang6beyond}, the aperture of the cone was used to study frequent and less frequent regions of interest.

\subsection{Modelling light in indoor environments}
In previous studies, indoor light modeling is mostly a field of research in visual computing and computer graphics. However, in these fields more emphasis is given on the generation of lifelike and photorealistic renderings rather than the actual spatial lighting measurement. On the other hand in the lighting field the focus is given on commercial CAD-design modeling software products, \eg Relux \cite{relux}, DIALux \cite{dialux} and AGi32 \cite{agi32}, which are broadly used for offline measurement and evaluation of lighting solutions in a simulated environment. To the best of our knowledge, the RGBD2Lux approach \cite{tsesmelis2018} is the first to bridge the gap across the two fields, bringing together visual computing and lighting design. By using only RGBD input, the method obtains a dense light intensity estimation of an indoor environment.

% as in lighting design software, we target per-pixel exact lighting intensity estimates using a radiosity model that accounts for realistic environments. By contrast, computer graphics approaches pose emphasis on photorealistic appearances, caring more for shadows, reflections, liquids and smoke, rather than the actual lighting measure as we are interested in this application.

% Relux \cite{relux}, DIALux \cite{dialux} and AGi32 \cite{agi32} are commercial CAD-design modeling software products, and commonly used in the lighting design field, for measurement and evaluation of lighting solutions. However, their main target is the evaluation of a lighting system installation rather than modelling for productivity or power efficiency (though this aspect is addressed to some level). 

% Moreover, their main constraint of providing all the needed information in advanced was only recently tackled by Tsesmelis \etal \todo{add ref} where he presents an on the fly light estimation procedure where only the light source properties (location and intensity) need to be provided in advance.

% \begin{itemize}
%     \item Introduction of people detection and pose estimation, specific reference to top-view approaches.
%     \item Introduction of light estimation
%     \item ego-centric object perception and activity estimation. Here we make it for ego-light-perception.
% \end{itemize}

\section{Ego-light-perception}\label{sec:method}

Any light management system that has to autonomously adjust the illumination of the environment  %based on some predefined lighting scenarios or lighting ISO standards targeting either productivity or power efficiency or even both,
has to be aware of two main factors: the human occupancy and their activity in the environment (human centric analysis) and the existing ambient illumination over time considering how is this influenced from the scene structure, the object materials and the light sources (scene composition analysis).

These two aspects are tightly intertwined, since the structure of the scene allows and constrains human activities, but at the same time the human activities influence the scene structure. Consider for example a warehouse as a scene: its structure continuously changes due to the different arrangement of the goods, the latter being a direct consequence of the human activities carried out in the environment. In other words, the structure of the scene and the human have to be considered as parts of a whole, accounting in addition for their continued temporal evolution.

To this end, the major goal in this work is to provide a new computer vision system for estimating the illumination map along with the human occupancy and attention from a single view.
We do this by bringing together individual works into a unique pipeline as we show in Figure \ref{fig:teaser}. 
% FIXME: here you could continue to describe the system if there is no overlap with previous text.

\subsection{People detection and head-pose estimation} \label{sec:head_detection_orientation}

We aim to detect people and estimate their head pose (their viewing angle). For the first task we adapt the Mask R-CNN \cite{he2017mask} object detector, while for the second one the head pose estimator proposed in \cite{hasan2017tiny}.  

The R-CNN \cite{he2017mask} detector has the ResNet-101 \cite{he2016deep} as a backbone architecture, trained on 80k images and 35k subset of evaluation images (trainval35k) of MS COCO dataset \cite{lin2014microsoft}.  
%%MARCO: here above add citations
We fine-tune the detector on our top-view dataset (see Sec. \ref{sec:dataset}), adopting a specific training portion of the data. We randomly partition the data into training and testing set, keeping 70\% of the data for training and 30\% for testing. 
%%MARCO: you need to specify this data
%We keep the alternative training approach of Mask RCNN\cite{he2017mask} as illustrated in the paper. 
Since the top-view images are different from the frontal-view images of the  COCO dataset \cite{lin2014microsoft}, the fine-tuning has a crucial role. We adopt a similar procedure for training the head pose estimator as in \cite{hasan2017tiny}. It is worth noting that the input for the head pose is the whole body detection bounding box: this is because \cite{hasan2017tiny} has been specifically designed for managing small-sized head patches, exploiting the body as contextual cue for a better final head orientation classification. In particular, 4 and 8 classes related to angles have been taken into account. 

During testing time, a cascaded approach is followed, first by applying the people detector and then feeding the detected body bounding box as input into the head orientation module.
\iffalse
We use a pre-trained network for head pose estimation as well and subsequently fine tune the last fully connected layers on our dataset. The output of the head detector is the class label (orientations are quantize into K-classes). During the inference time, we have a cascaded approach. First, we detect people and then these detections are used as input to the head pose estimator. We quantize the orientation of the person into four classes and output of the head pose estimator is the softmax over K possibilities. Where K is number of classes. 
\fi

% \todo{ please adapt it to the current introduction and to the needed parts, trying to include into the storyline what we have done, eg forecast.
% \begin{itemize}
%     \item detect people
%     \item estimate their head orientation
%     \item forecast their movements (directly regressing point instead of sampling from distribution)
%     \item forecast their head orientation
% \end{itemize}}

\subsection{Spatial light estimation}
To obtain an estimate of a dense spatial illumination map, we adapt our work in \cite{tsesmelis2018}. In this work we make use of a radiosity model \cite{cohen1993rri} for estimating the spatial illumination over time by just using the input from an RGBD camera. Furthermore, we extract the information regarding the photometric properties of the material of the scene based on a photometric stereo baseline approach that is applied on the time-varying RGB images. This approach extracts a scalar albedo at each pixel by using a set of images with different light sources that are switched on/off during the day.  % while at the same time they compose the partial geometry of the scene from a depth sensor. 
Having the light sources position and intensity, the scalar albedo under Lambertian assumptions, and the depth map from the sensor, our proposed method in \cite{tsesmelis2018} shows that it is possible to obtain a dense measurement of the light emitted by a 3D patch in the indoor environment.
In order to provide more realistic estimates, we model real lighting systems that, differently from point-like sources, emit light given a specific light distribution curve (LDC). The LDC is custom for each lighting system and their properties are considered to be known when estimating the light instensity. %which considered to be the only known parameter in the radiosity equation and provide a radiosity solution for each patch to which the scene is discretized. 
The proposed method shows that, even by accounting the non-linearities of LDC, it is possible to solve for the radiosity equation with Least Squares and so obtain a more reliable measure of the light intensity. %The radiosity solution is then considered as the spatial illumination estimation solution based on the Lambertian assumption. 
For the evaluation of our approach we used point-to-point sensory equipment \aka luxmeters installed across the scene.

\subsection{Gaze-gathered light modelling}
%As it is mentioned m

\begin{figure}[thbp]
\vspace{-10pt}
	\begin{center}

		\subfloat[][]{\includegraphics[width=0.45\linewidth]{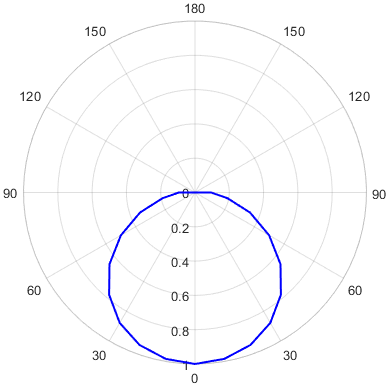}\label{fig:lsc}}
		\hspace{0.01em}
		\subfloat[][]{\includegraphics[width=0.45\linewidth]{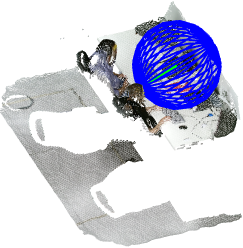}\label{fig:lsc_modeling}}
% 		\hspace{0.01em}
% 		\subfloat[][]
% 		{\includegraphics[width=0.32\linewidth]{images/unit_sphere_over_cad.png}\label{fig:isocel_rays}}
        \vspace{5pt}
		\caption{Modeling of the Luxmeter Sensitivity Curve (LSC) as a human light perception model.}
		\label{fig:lsc_sampling}
	\end{center}
	\vspace{-10pt}
\end{figure}

Light measurements are practically made using a luxmeter sensor. This sensor measures the perceived light that is in function of the distance to the light, the orientation and other manufacturing characteristics. These properties are resumed by the Luxmeter Sensitivity Curve (LSC) as in Figure \ref{fig:lsc}.
%
%Measuring human light perception is not a trivial procedure and conclusions are hard to draw in every scenario \cite{iesna2000,bernecker2017}. People perceive light differently, depending on their orientation with respect to the light sources (whether they are frontally illuminated or from the side) and depending on the distance from it (FIXME: this is a bit dangerous, why you do not say that luxmeter have encoded a curve that model perception? So we put all the perceptual problems aside, and people can blame luxmeters, not us). 
%In a similar fashion, luxmeter sensors have different sensitivity to lighting, depending on the lighting angle and distance as well as to and to the manufacturing characteristics. The LSC curve in Figure \ref{fig:lsc} 
The LSC illustrates the  perception characteristic of every luxmeter sensor which in this work we adopt in order to meet the measuring requirements of the collected ground truth data and to simulate the human light perception. We have chosen this solution because this is the standard de facto in the lighting industry and it provides satisfactory solutions when doing light commissioning \cite{ies2011commissioning}.

% from the human perspective (FIXME: watch out again...). We ended with the solution of using a luxmeter since it is the only simple and practical way in comparison to a perceptual study .

The key idea in this procedure is that, once we have detected a person in the image and estimated his head positioning and orientation as described in Sec. \ref{sec:head_detection_orientation}, we extract his posture in the 3D space by mapping the 2D image coordinates of his detected head to the corresponding depth information. Thereafter, once we have the positioning of the head in the 3D space as well as its orientation (where the person looks at), we estimate the light that arrives to his/her face (or to the luxmeter as in our case) by applying a ray-casting procedure where we simulate the human field of view (FOV). Such view frustum is obtained by using emitted rays starting from the estimated head position towards the corresponding estimated head orientation. The total illumination arriving to the person is computed by adding the related spatial illumination (radiance) from the patches of the scene that are in the direct visibility of the person. The rays project in the space as a uniform generated sequence over the unit sphere and weighted accordingly, based on the modelled luxmeter's LSC, towards the visible patches from the FOV of the sensor. % and how the sensor perceives light (stronger in the center, weaker in the sides following the human sight). 
The contribution of each patch to the total amount of lighting perceived by the occupant, is computed by estimating the percentage of rays intersecting that patch.

% Finally the fraction of the amount of rays intersecting the visible patches over the total emitted patches from the initial patch corresponding to the head positioning provides us with the contribution of each patch to the total amount of light arriving to the occupants (FIXME rephrase).

\section{Results}\label{sec:evaluation}
Experiments are organized as follows, Sec. \ref{sec:dataset} presents the recorded dataset with all the different ablation studies both for light measurements as well as for top-view detection and head-pose estimation. Sec. \ref{sec:head_estimation_eval} reports the results regarding the person occupancy and head pose estimation study, while Sec. \ref{sec:light_perception} describes in more details the evaluation for both spatial and the gaze-gathered light estimation. Finally, Sec. \ref{sec:applications} evaluates the Invisible Light Switch as a power saving application.

\subsection{Dataset overview}\label{sec:dataset}
\begin{figure}[thbp]
\vspace{-5pt}
	\begin{center}

		\subfloat[][]{\includegraphics[width=0.49\linewidth]{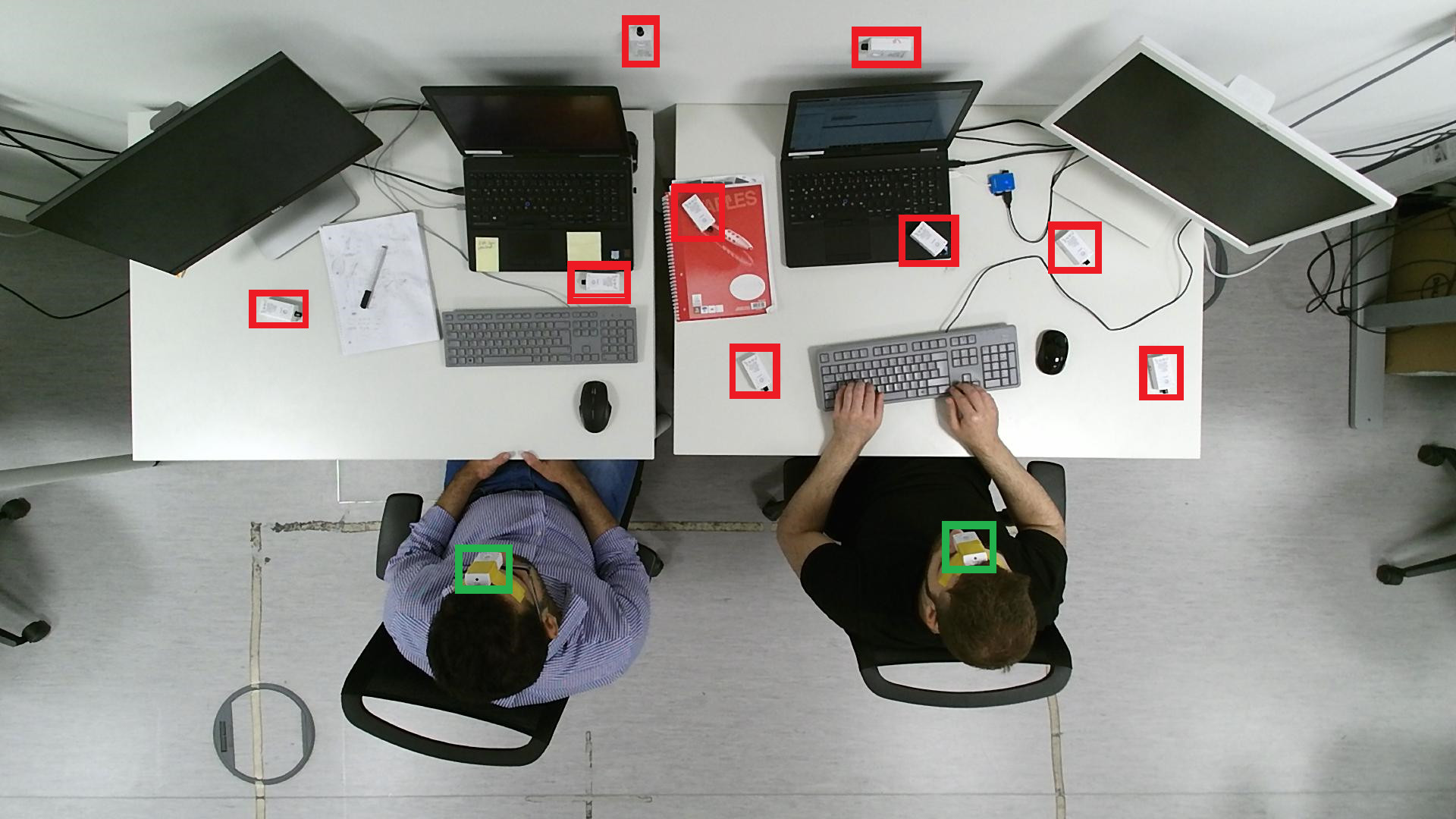}\label{fig:room6}}
		\hspace{0.01em}
		\subfloat[][]{\includegraphics[width=0.49\linewidth]{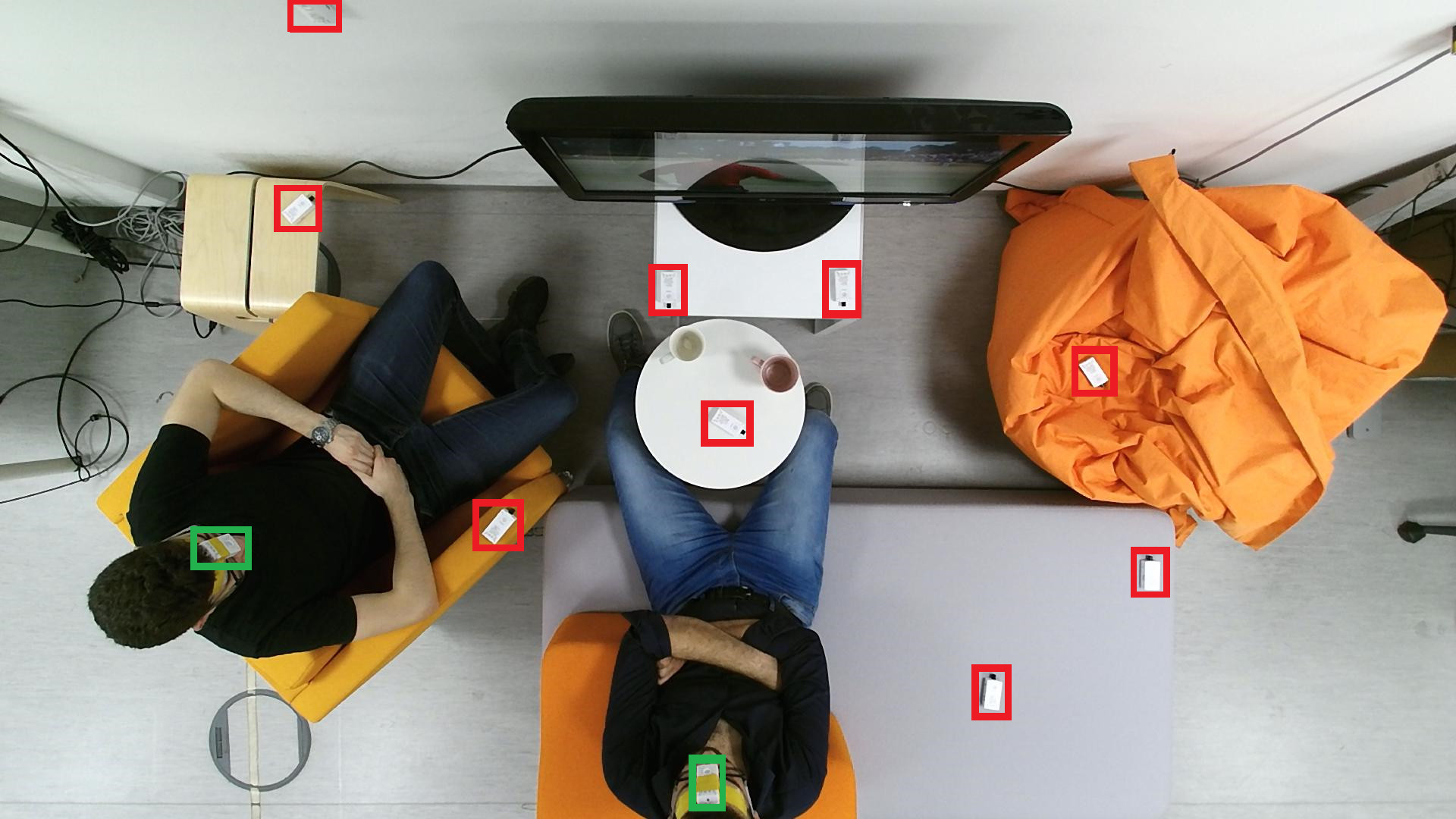}\label{fig:room7}}
% 		\hspace{0.01em}
% 		\subfloat[][]
% 		{\includegraphics[width=0.32\linewidth]{images/unit_sphere_over_cad.png}\label{fig:isocel_rays}}
        \vspace{5pt}
		\caption{Illustration of the two indoor scenes used for evaluation: (a) illustrates a normal office environment and (b) shows a relaxing area. Red and green bounding boxes are showing the location of luxmeters within the space covering the spatial and gaze-gathered illumination ground truth measurements respectively.}
		\label{fig:new_scenes}
	\end{center}
	\vspace{-10pt}
\end{figure}

To the best of our knowledge, Tsesmelis' \etal work \cite{tsesmelis2018} is the first to introduce a dataset for benchmarking light measurements with ground truth sensory data in real scenes. In this work we extended this dataset by introducing two more scenes with human activity, one based on a normal office environment and a second one representing a relaxing area (see Figure \ref{fig:new_scenes}). 

Both scenes comprehend different human activities \eg watching TV, working on a desk area, chatting, \etc, as well as different head orientations (VFOA) and multiple light combinations. In this work, VFOA is a cone with vertex in the middle of a person's eyes,  oriented as the gaze direction and an aperture angle of  $\alpha = \ang{30}$.

\begin{figure}[ht!]
	\begin{center}
		\includegraphics[width=1\linewidth]{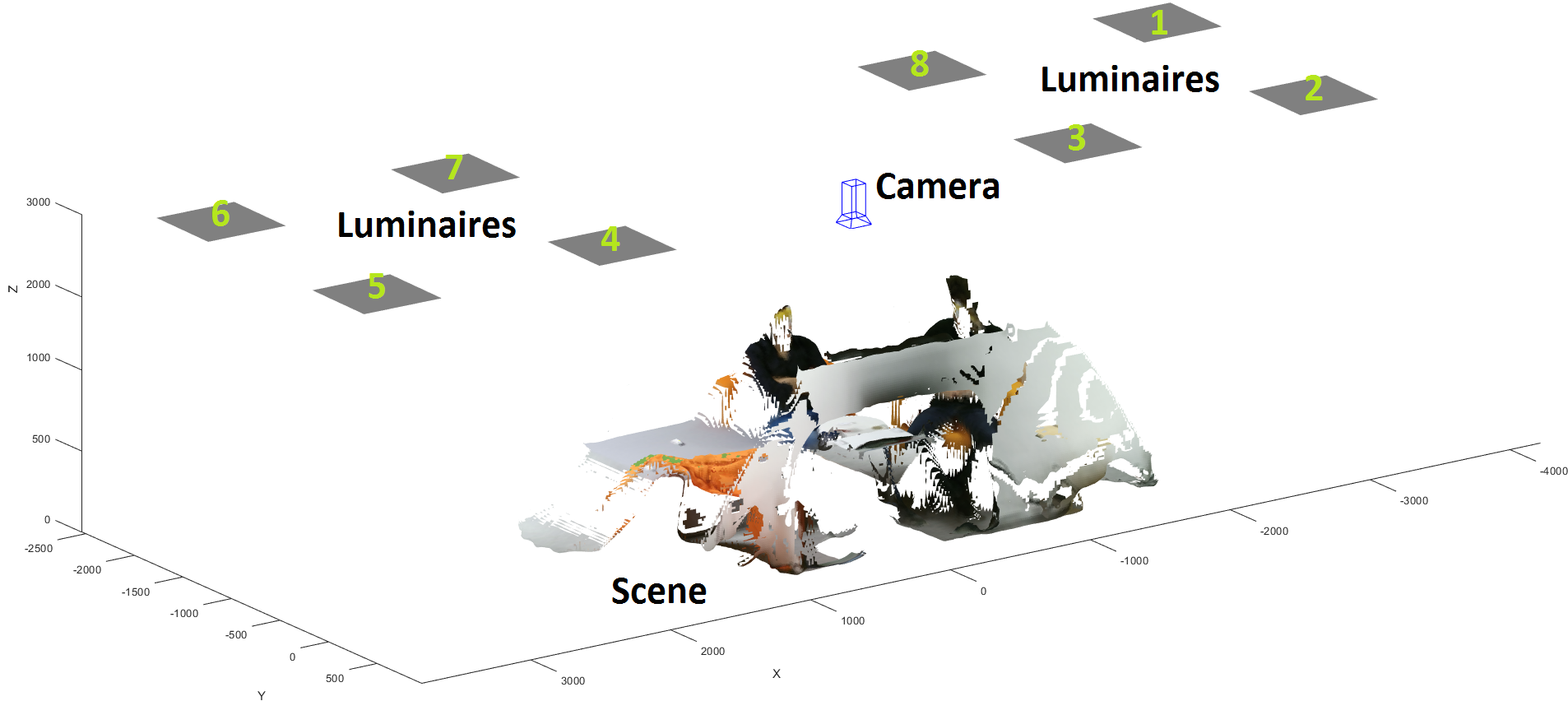}
	\end{center}
	\caption{Illustration of the light management installation.}\label{fig:luminaires}
	\vspace{-4pt}
\end{figure}

In both rooms there is a controlled light management installation, where the position, type and properties (\eg luminous intensity, light distribution curve, \etc) of the luminaires (eight in total) are considered known, see Figure \ref{fig:luminaires}.

For obtaining the ground truth data we have installed and used a number of sensory equipment. A calibrated and aligned RGBD camera system (Kinect v2) is installed in the ceiling of the room providing a top-view perspective of the scene, see Fig. \ref{fig:new_scenes} and \ref{fig:luminaires}. Moreover, the camera is synchronized with a number of luxmeters (also indicated in Fig. \ref{fig:new_scenes}) providing the light intensity ground truth data both for the spatial as well as for the gaze-gathered (attached to the forehead of the occupants) illumination. Considering the limitation (\ie point-to-point) of lux readings that the luxmeters provide, we installed $11$ sensors in different areas, thus providing a reasonable sampling of the scene. We use $9$ luxemetes for evaluating the spatial illumination across the environment and $2$ luxmeters for measuring the light intensity that arrives to each one of the occupants appearing in the scenes. For each luxmeter, we additionally report the type and their specific light sensitivity characteristic curve, LSC (see Fig. \ref{fig:lsc_sampling}) giving the sensor's sensitivity across the incident light angles.

\begin{figure}[ht!]
	\begin{center}
		\includegraphics[width=1\linewidth]{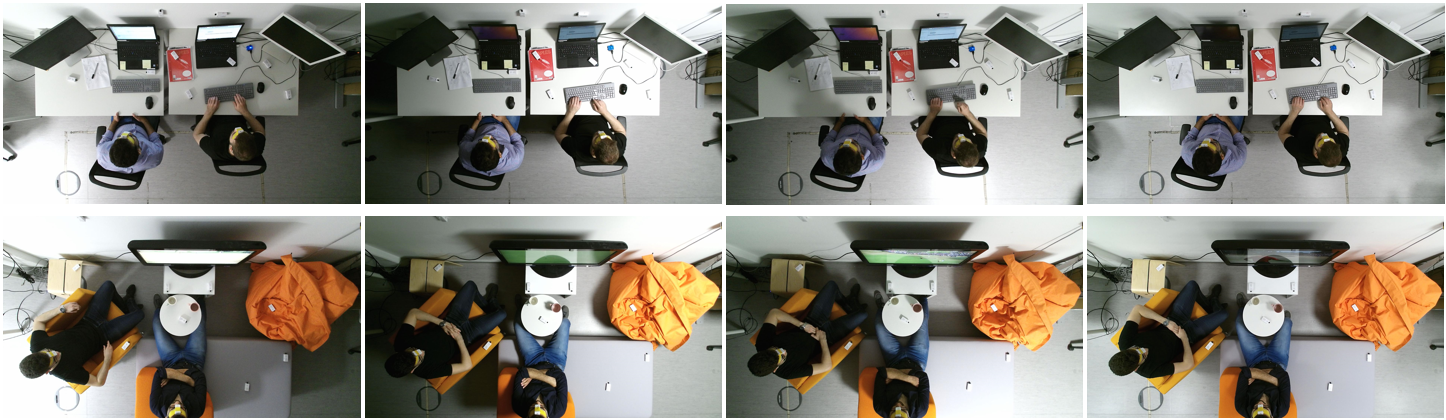}
	\end{center}
	\caption{Illustration of 4 illumination variants within the two rooms. From left to right, the images illustrate the illumination provided by 1, 4, 7 and all 8 luminaires switched on in the two scenes.}\label{fig:illumination_combinations}
	\vspace{-5pt}
\end{figure}

Thereafter, we evaluate $24$ and $30$ different scenarios with different luminaire activations (luminaires switched on/off) for each room respectively (see Fig. \ref{fig:illumination_combinations}). We target the use of RGB and depth input just for light measurement, the use of luxmeters as ground truth, and all other provided information for evaluation studies.

\subsection{Top-view detection and head-pose estimation}\label{sec:head_estimation_eval}
% In the recorded dataset we consider 5 and 4 different \textbf{Ab}lation \textbf{S}tudies (\aka AbS) for each one of the scenes respectively based on the visual attention of the individuals. For the office scene we have the following use cases:\\
%     \textbf{AbS 1:} both looking straight to their monitor\\
%     \textbf{AbS 2:} both looking to the left monitor\\
%     \textbf{AbS 3:} both looking to the right monitor\\
%     \textbf{AbS 4:} both looking to each other\\
%     \textbf{AbS 5:} both looking to the right\\

% For the resting area accordingly we have:\\
%     \textbf{AbS 1:} both looking to the TV\\
%     \textbf{AbS 2:} both looking to the right\\
%     \textbf{AbS 3:} both looking to each other\\
%     \textbf{AbS 4:} one looking to the TV, while the other to him\\

\begin{figure}[!ht]
	\begin{center}
	    \includegraphics[width=0.85\linewidth]{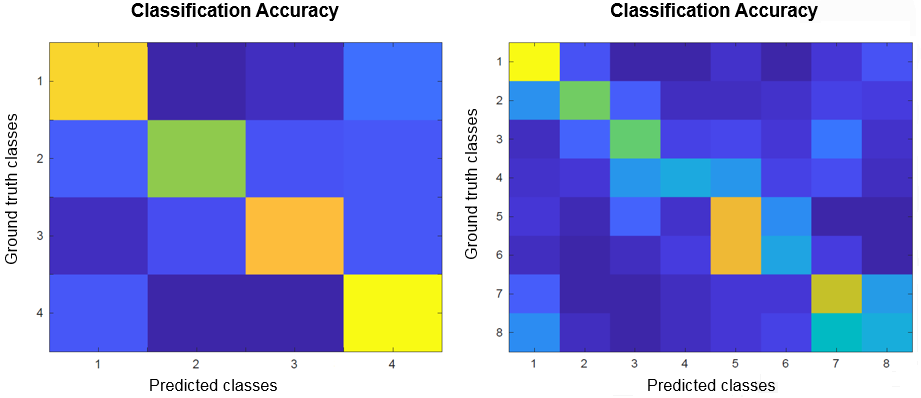}
	\end{center}
	\caption{Confusion matrices of the head pose estimator. From left to right, the 4 and 8 classes confusion matrix respectively.}\label{fig:confMat}
%  	\vspace{-5pt}
\end{figure}

% \todo{Irtiza please start adding images and description
% \begin{itemize}
%     \item Trajectory forecasting evaluation (MAD and FAD)
%     \item Head orientation results (MA)
%     \item \textbf{Optional} person detection evaluation (LAMR or AP)
% \end{itemize} }
We fine tune both the person detector and the head pose estimator on our top-view dataset. We report an average precision (AP) of 98\% in terms of people detection. As mentioned previously we test our approach on the testing set of our top-view dataset.
%%MARCO you need to specify the testing data
For the head pose orientation fine tuning on the whole body has been crucial for the performance, since using the sole head region produced definitely worst scores. In particular, 
%However, head pose estimation in top view images is a challenging task especially when images are acquired from far, such as in our top view dataset. Due to the small head region, during our experiments we found it hard to train the network and achieve high accuracy. Therefore we adopted the approach proposed in \cite{hasan2017tiny}, where input to the head pose estimator is the complete body of an individual detected by Mask R-CNN as illustrated in \ref{fig:fulldetpip}. 
we adopt two different class numbers for head pose, namely 4 and 8. The corresponding confusion matrices are reported in Fig. \ref{fig:confMat}, showing an accuracy of 43.2\% (8 classes) and 70.7\% (4 classes) respectively. The scarce performance in the 8-class case is due to the mix among adjacent viewing angles: actually, the average size of the head region in the dataset is approx. 40x50 pixels.
%%MARCO put the average head region size here above
\iffalse
, due to lower inter-class variation majority of miss classifications for a given class label are observed in previous or next classes. Given a class label of 5 (8 classes, right image) the class label 5 gets wrongly classified to the next class label 6. Therefore, in order to increase the inter class variation, we quantify the head orientation in 4 discrete classes. Intuitively, the classification accuracy of the of head pose estimator increased from 43.2\%(8 classes) to 70.7\%(4 classes). 
\fi 
For these reasons, we use the 4-class version in the light perception studies.        

\begin{figure}[thbp]\label{fig:fulldetpip}
	\begin{center}

		\subfloat[][]{\includegraphics[width=0.49\linewidth]{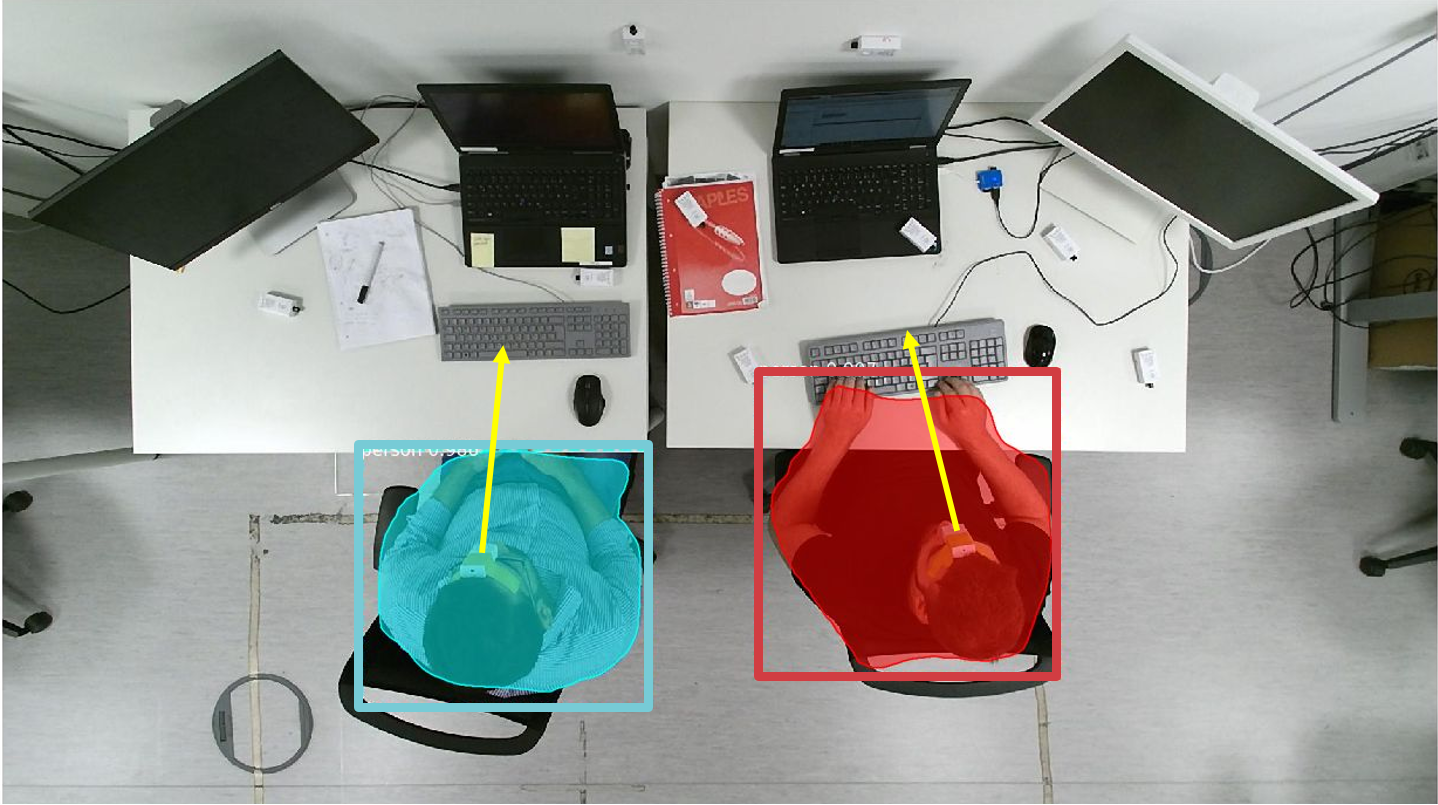}\label{fig:det-hpe}}
		\hspace{0.01em}
		\subfloat[][]{\includegraphics[width=0.49\linewidth]{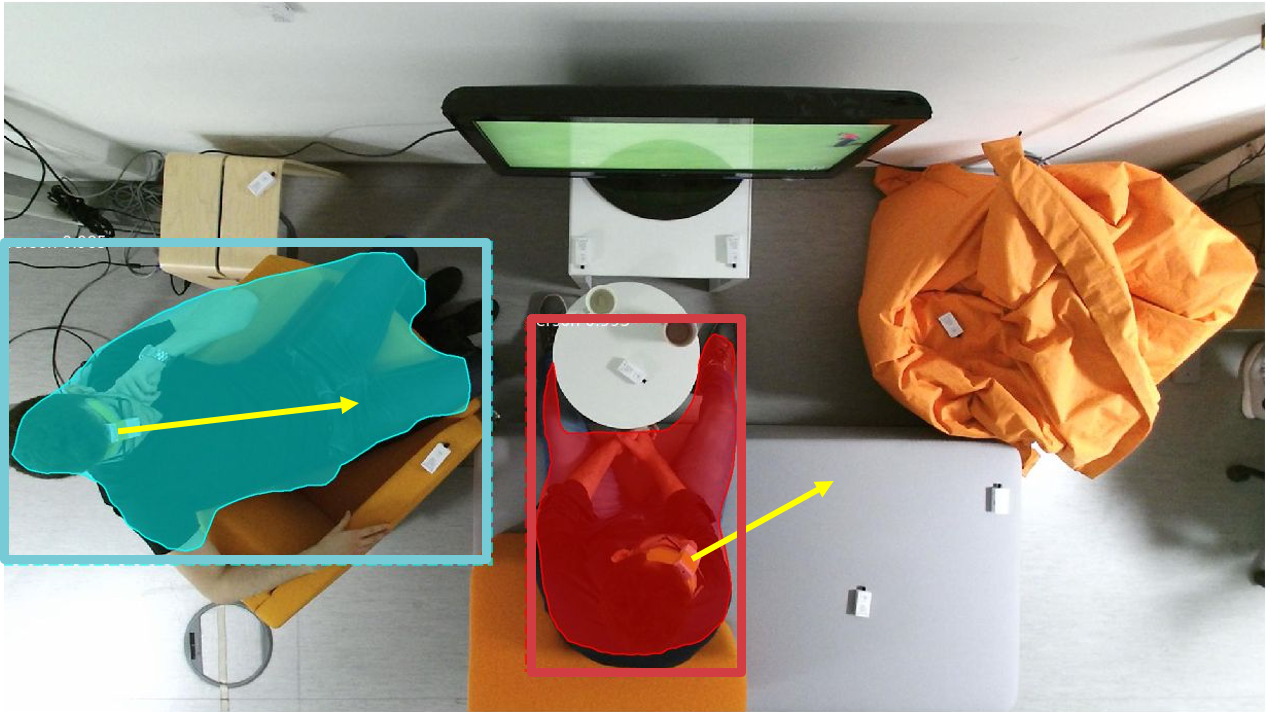}\label{fig:det-hpe-2}}
% 		\hspace{0.01em}
% 		\subfloat[][]
% 		{\includegraphics[width=0.32\linewidth]{images/unit_sphere_over_cad.png}\label{fig:isocel_rays}}
        \vspace{5pt}
		\caption{Illustration of people detection and head pose estimation. We detect people in the scene by using Mask R-CNN and then the detections are provided as input to the head pose estimator.}
		\label{fig:human_detector}
	\end{center}
	\vspace{-10pt}
\end{figure}

% \begin{figure}[thbp]
% 	\begin{center}

% 		\subfloat[][]{\includegraphics[width=0.49\linewidth]{images/confMat_4.png}\label{fig:det-hpe}}
% 		\hspace{0.01em}
% 		\subfloat[][]{\includegraphics[width=0.49\linewidth]{images/confMat.png}\label{fig:det-hpe-2}}
% % 		\hspace{0.01em}
% % 		\subfloat[][]
% % 		{\includegraphics[width=0.32\linewidth]{images/unit_sphere_over_cad.png}\label{fig:isocel_rays}}
%         \vspace{5pt}
% 		\caption{head orientation results 4 v 8}
% 		\label{fig:new_scenes}
% 	\end{center}
% 	\vspace{-10pt}
% \end{figure}

% \usepackage{multirow}
\begin{table}[!ht]
\centering
{\small
\caption{ The values represent the average estimated illumination error over the different lighting activation \wrt the ground truth measurements, for both scenes. Columns 1-9 corresponds to the spatial average values for the corresponding installed luxmeters in the environment. By contrast, values in columns 10-11 consider those luxmeters for evaluating the human light perception.}}
\label{table:quantitative_scene1_2}
\resizebox{.48\textwidth}{!}{\begin{tabular}{ccccccccccccccc}
\hline
\multicolumn{2}{|c|}{\multirow{3}{*}{}}                                                                                                   & \multicolumn{13}{c|}{\textbf{\begin{tabular}[c]{@{}c@{}}Avg. error $\varepsilon$\\ (in Lux)\end{tabular}}}                                                                                                                                                                                                                                                                                                                                                                                                                                                                    \\ \cline{3-15} 
\multicolumn{2}{|c|}{}                                                                                                                    & \multicolumn{11}{c!{\vrule width2pt}}{\textbf{Luxmeters}}                                                                                                                                                                                                                                                                                                                                              & \multicolumn{2}{l|}{}                                                                                                                                                     \\ \cline{3-15} 
\multicolumn{2}{|c|}{}                                                                                                                    & \multicolumn{1}{c|}{\textbf{1}} & \multicolumn{1}{c|}{\textbf{2}} & \multicolumn{1}{c|}{\textbf{3}} & \multicolumn{1}{c|}{\textbf{4}} & \multicolumn{1}{c|}{\textbf{5}} & \multicolumn{1}{c|}{\textbf{6}} & \multicolumn{1}{c|}{\textbf{7}} & \multicolumn{1}{c|}{\textbf{8}} & \multicolumn{1}{c|}{\textbf{9}} & \multicolumn{1}{c|}{\textbf{10}} & \multicolumn{1}{c!{\vrule width2pt}}{\textbf{11}} & \multicolumn{1}{c|}{\textbf{\begin{tabular}[c]{@{}c@{}}Avg.\\ (1-9)\end{tabular}}} & \multicolumn{1}{c|}{\textbf{\begin{tabular}[c]{@{}c@{}}Avg.\\ (11-10)\end{tabular}}} \\ \hline
\multicolumn{1}{|c|}{\multirow{2}{*}{\textbf{Scene 1}}} & \multicolumn{1}{c|}{\begin{tabular}[c]{@{}c@{}}$\varepsilon_{est}$\\ (\wrt GT)\end{tabular}}    & \multicolumn{1}{c|}{62.5}       & \multicolumn{1}{c|}{26.3}       & \multicolumn{1}{c|}{68.0}       & \multicolumn{1}{c|}{65.1}       & \multicolumn{1}{c|}{47.9}       & \multicolumn{1}{c|}{57.1}       & \multicolumn{1}{c|}{44.0}       & \multicolumn{1}{c|}{29.9}       & \multicolumn{1}{c|}{28.0}       & \multicolumn{1}{c|}{97.6}        & \multicolumn{1}{c!{\vrule width2pt}}{92.2}        & \multicolumn{1}{c|}{56.2}                                                          & \multicolumn{1}{c|}{94.7}                                                            \\ \cline{2-15} 
\multicolumn{1}{|c|}{}                                  & \multicolumn{1}{c|}{\begin{tabular}[c]{@{}c@{}}$\varepsilon_{est\_d}$\\ (\wrt GT)\end{tabular}} & \multicolumn{1}{c|}{-}          & \multicolumn{1}{c|}{-}          & \multicolumn{1}{c|}{-}          & \multicolumn{1}{c|}{-}          & \multicolumn{1}{c|}{-}          & \multicolumn{1}{c|}{-}          & \multicolumn{1}{c|}{-}          & \multicolumn{1}{c|}{-}          & \multicolumn{1}{c|}{-}          & \multicolumn{1}{c|}{216.08}            & \multicolumn{1}{c!{\vrule width2pt}}{166.4}            & \multicolumn{1}{c|}{-}                                                             & \multicolumn{1}{c|}{191.24}                                                                \\ \hline
\textbf{}                                               &                                                                                 &                                 &                                 &                                 &                                 &                                 &                                 &                                 &                                 &                                 &                                  &                                  &                                                                                    &                                                                                      \\ \hline
\multicolumn{1}{|c|}{\multirow{2}{*}{\textbf{Scene 2}}} & \multicolumn{1}{c|}{\begin{tabular}[c]{@{}c@{}}$\varepsilon_{est}$\\ (\wrt GT)\end{tabular}}    & \multicolumn{1}{c|}{35.3}       & \multicolumn{1}{c|}{33.8}       & \multicolumn{1}{c|}{44.0}       & \multicolumn{1}{c|}{20.1}       & \multicolumn{1}{c|}{31.5}       & \multicolumn{1}{c|}{39.6}       & \multicolumn{1}{c|}{23.6}       & \multicolumn{1}{c|}{27.9}       & \multicolumn{1}{c|}{27.3}       & \multicolumn{1}{c|}{41.7}        & \multicolumn{1}{c!{\vrule width2pt}}{69.2}        & \multicolumn{1}{c|}{35.8}                                                          & \multicolumn{1}{c|}{55.4}                                                            \\ \cline{2-15} 
\multicolumn{1}{|c|}{}                                  & \multicolumn{1}{c|}{\begin{tabular}[c]{@{}c@{}}$\varepsilon_{est\_d}$\\ (\wrt GT)\end{tabular}} & \multicolumn{1}{c|}{-}          & \multicolumn{1}{c|}{-}          & \multicolumn{1}{c|}{-}          & \multicolumn{1}{c|}{-}          & \multicolumn{1}{c|}{-}          & \multicolumn{1}{c|}{-}          & \multicolumn{1}{c|}{-}          & \multicolumn{1}{c|}{-}          & \multicolumn{1}{c|}{-}          & \multicolumn{1}{c|}{55.42}            & \multicolumn{1}{c!{\vrule width2pt}}{151.93}            & \multicolumn{1}{c|}{-}                                                             & \multicolumn{1}{c|}{103.68}                                                                \\ \hline
\end{tabular}}
\end{table}

\subsection{Person-perceived light estimation} \label{sec:light_perception}

Table \ref{table:quantitative_scene1_2} presents the quantitative results of our adopted light estimation approach. The table shows the average estimated error in lux values for both spatial (luxmeters 1-9) and gaze-gathered light estimation (luxmeters 10-11) cases. It can be easily noticed that the error, $\varepsilon_{est}$, for all luxmeters does not exceed the range of 100 lux, this yields an overall average light estimation error approx. 56 lux for Scene 1 and 36 lux for Scene 2. On the other hand, if we now consider only the luxmeters intended for evaluating the gaze-gathered light estimation, \ie luxmeters 10 and 11, we notice that the error raises up to 94.7 lux and 55.4 lux for each scene respectively. This can be justified due to inaccuracies in the reconstruction of the 3D mesh areas corresponding to the head position and orientation of the occupants, as well as to the fact that the inter-reflections from the wall towards the sensors are limited due to incomplete reconstruction as an outcome of the limited FOV of the depth sensor. In any case, the fact that the average light estimation error does not exceed 100 lux indicates that the estimated illumination map can be considered reliable for describing the global illumination of the scene.

% \begin{figure}[!ht]
%   \begin{center}
%     \includegraphics[width=0.48\linewidth]{images/wrongheadpose.png}
%   \end{center}
%   \caption{Birds}
% \end{figure}

Furthermore, to demonstrate the applicability of our model, we use as explained a real person detector and a head pose estimator (making the pipeline completely automatic). In Table \ref{table:quantitative_scene1_2} the $\varepsilon_{est\_d}$ rows for column 10 and 11, illustrates the error based on the detectors output for both scene 1 and 2. It can be observed that while the average error \wrt the oracle is less than 100 lux, this error raises up to the range of 200 lux negative variation \wrt to the ground truth measurements. The last can be justified by erroneous head pose estimations, considering the large step size (\ang{90}) of the 4-class adapted classification problem.  This further brings into discussion the fact that this error could further be substantially reduced by improving the head pose estimator.  

\begin{figure}[!ht]
	\begin{center}
	    \includegraphics[width=1\linewidth]{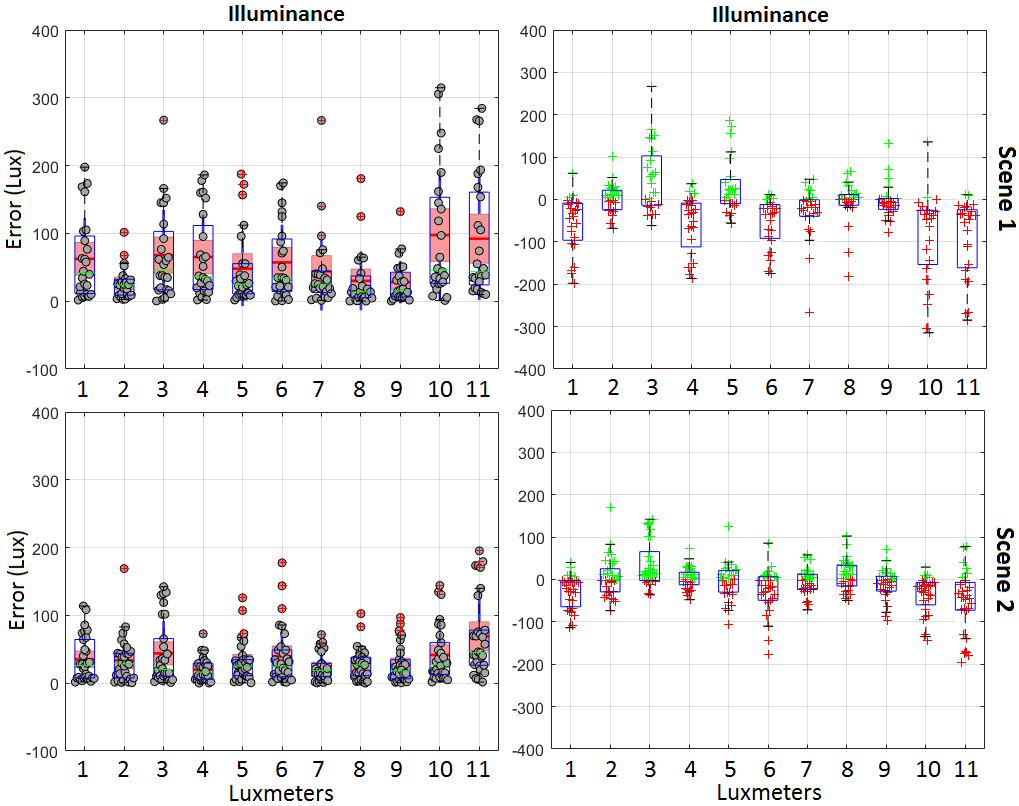} %\hspace{.3cm}
	\end{center}
	\caption{Scene 1 \& 2 boxplot error evaluation (in Lux) using based on the presented framework. The boxplots in the first and second columns show the absolute and signed illumination estimated error for each lighting scenario in each scene respectively.} \label{fig:rooms_full_scene_neg}
% 	\vspace{-5pt}
\end{figure}

Figure \ref{fig:rooms_full_scene_neg} shows in a graph analysis the values presented in Table \ref{table:quantitative_scene1_2}. The left graphs show the absolute light estimation error (y-axis), as estimated for each of the $11$ (9 for spatial and 2 for the human light perception) used luxmeter sensors (x-axis). The gray dots, forming each of the box plot boxes, represent the estimated error of each of the lighting scenarios for each scene while the pink box represents the central 50\% of the data. The upper and lower vertical lines indicate the extension of the remaining error points outside it and the central red line indicates the mean error which comes in alignment with the values shown in Table \ref{table:quantitative_scene1_2}. Similarly, the boxplots on the right present the signed illumination error accordingly. The green and red markers indicate whether the error is due to an over or under estimation of the illuminance at the sensor's location respectively. As it can be noticed in the most of the cases the error is a result of an under estimation of the illuminance which as explained earlier are a cause of the incomplete geometry of the scenes as we only consider the parts of the environment within the FOV of the camera sensors.

\begin{figure}[!ht]
	\begin{center}
	    \includegraphics[width=1\linewidth]{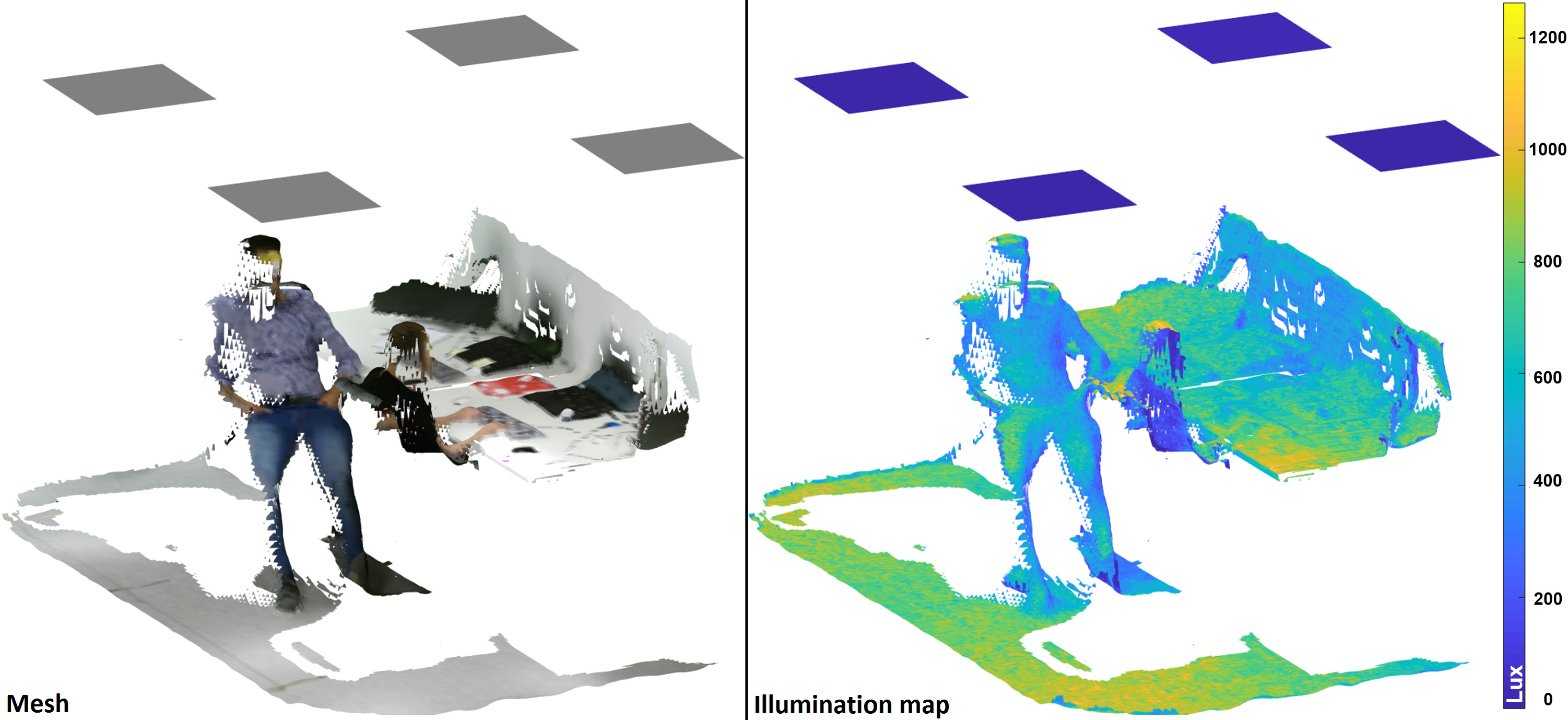} %\hspace{.3cm}
	\end{center}
	\caption{Illumination map of the full-lit scenario in scene 1 with a dense representation of the global illumination of the environment.} \label{fig:illuminance_map1}
% 	\vspace{-1pt}
\end{figure}

\begin{figure}[!htb]
	\begin{center}
	    \includegraphics[width=1\linewidth]{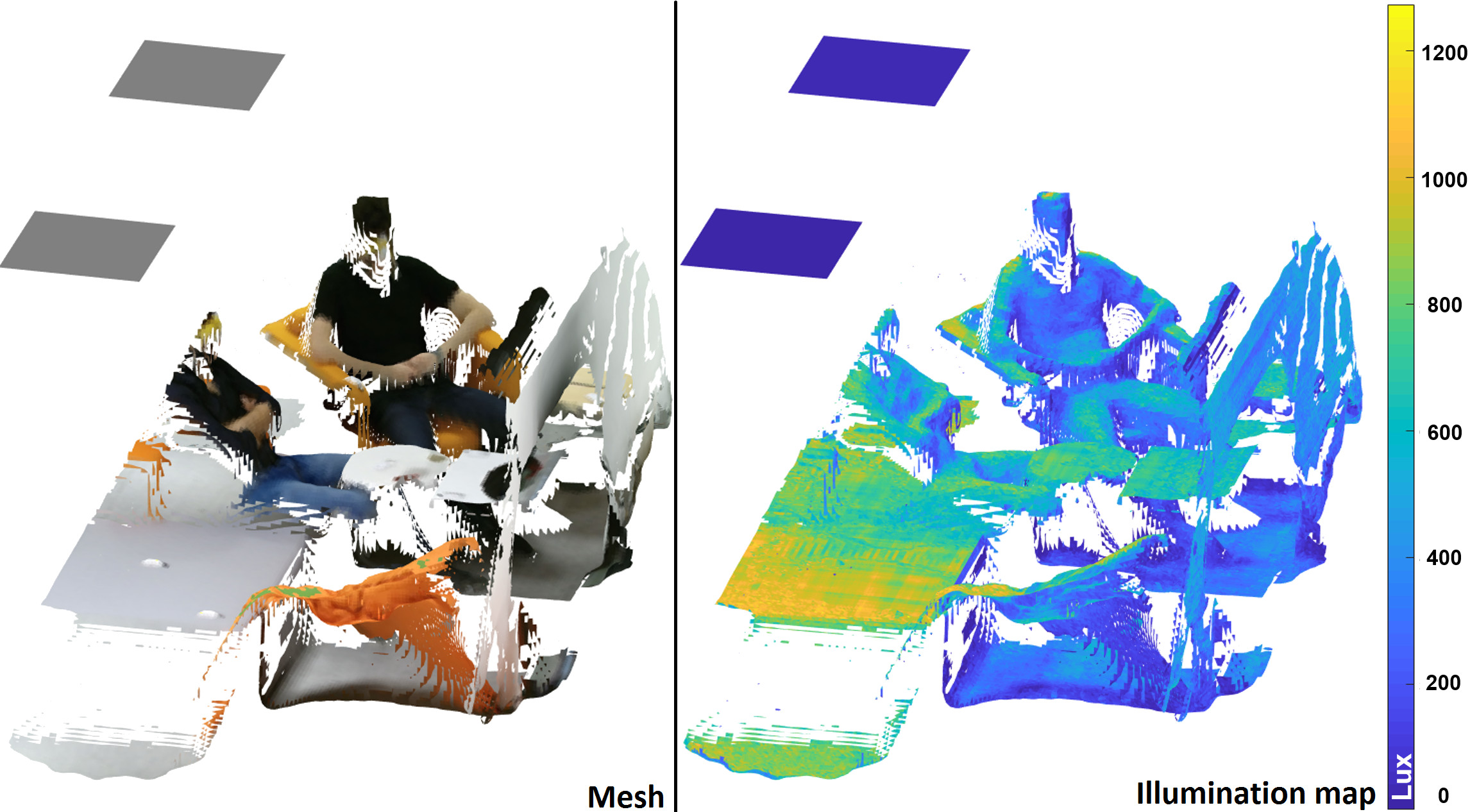} %\hspace{.3cm}
	\end{center}
	\caption{{\small Illumination map of the full-lit scenario in scene 2. Notice the estimated illumination in the area in front of the occupants which is less bright in comparison to the one that are on their side. This is due to the body occlusion on the direct illumination coming from the luminaires from their back which is correctly estimated by the ILS.}} \label{fig:illuminance_map2}
	\vspace{-5pt}
\end{figure}

Finally, figures \ref{fig:illuminance_map1} and \ref{fig:illuminance_map2} visualise the illumination maps in the 3D space for one of the illumination scenarios in each of the scenes. As it can be seen the visualized illumination maps provide an accurate dense representation of the global illumination of the environment over time.

\begin{table*}[!ht]
\centering
{\small\caption{ Quantitative analysis of four different head orientation class studies (VFOA), two for each scene. $\varDelta_{lux}$ shows the discrepancy of different lighting scenarios \wrt the full lit scenario (reference). $\varepsilon_{est}$ shows the corresponding average error of the estimated light in regards to the ground truth lux measurements and $\varDelta_{watt}$ shows the discrepancy of the power consumption in watts considering the active/non active luminaires for each corresponding scenario.}}
\label{table:power_saving}
\resizebox{.999\textwidth}{!}{\begin{tabular}{ccccc|ccc|ccc|cccc}
\hline
\multicolumn{2}{|c|}{\multirow{2}{*}{\textbf{}}}                                                                                                                                                      & \multicolumn{6}{c|}{\textbf{Scene 1}}                                                                                                                                                       & \multicolumn{7}{c|}{\textbf{Scene 2}}                                                                                                                                                                                                                        \\ \cline{3-15} 
\multicolumn{2}{|c|}{}                                                                                                                                                                                & \multicolumn{3}{c|}{\textbf{VFOA 1}}                                                         & \multicolumn{3}{c|}{\textbf{VFOA 2}}                                                         & \multicolumn{3}{c|}{\textbf{VFOA 1}}                                                         & \multicolumn{4}{c|}{\textbf{VFOA 2}}                                                                                                                          \\ \hline
\multicolumn{2}{|c|}{\textbf{\begin{tabular}[c]{@{}c@{}}Luminaire\\ activations\end{tabular}}}                                                                                                        & \multicolumn{1}{c|}{\textbf{3$\mid$4$\mid$7$\mid$8}} & \multicolumn{1}{c|}{\textbf{2$\mid$3$\mid$4$\mid$5}} & \textbf{3$\mid$4} & \multicolumn{1}{c|}{\textbf{3$\mid$4$\mid$7$\mid$8}} & \multicolumn{1}{c|}{\textbf{2$\mid$3$\mid$4$\mid$5}} & \textbf{3$\mid$4} & \multicolumn{1}{c|}{\textbf{3$\mid$4$\mid$7$\mid$8}} & \multicolumn{1}{c|}{\textbf{2$\mid$3$\mid$4$\mid$5}} & \textbf{3$\mid$4} & \multicolumn{1}{c|}{\textbf{1$\mid$2$\mid$3$\mid$4$\mid$5$\mid$6}} & \multicolumn{1}{c|}{\textbf{2$\mid$3$\mid$4$\mid$5}} & \multicolumn{1}{c|}{\textbf{1$\mid$3$\mid$4$\mid$6}} & \multicolumn{1}{c|}{\textbf{3$\mid$4}} \\ \hline
\multicolumn{1}{|c|}{\multirow{3}{*}{\textbf{\begin{tabular}[c]{@{}c@{}}\\\\Luxmeter\\ 10\end{tabular}}}} & \multicolumn{1}{c|}{\textbf{\begin{tabular}[c]{@{}c@{}}$\varDelta_{lux}$\\ (\wrt full-lit)\end{tabular}}}  & \multicolumn{1}{c|}{116.15}           & \multicolumn{1}{c|}{123.77}           & 189.01       & \multicolumn{1}{c|}{85.4}             & \multicolumn{1}{c|}{123.8}            & 163.85       & \multicolumn{1}{c|}{84.23}            & \multicolumn{1}{c|}{93.69}            & 151.92       & \multicolumn{1}{c|}{106.52}               & \multicolumn{1}{c|}{148.12}           & \multicolumn{1}{c|}{157.07}           & \multicolumn{1}{c|}{191.15}       \\ \cline{2-15} 
\multicolumn{1}{|c|}{}                                                                                & \multicolumn{1}{c|}{\textbf{\begin{tabular}[c]{@{}c@{}}$\varepsilon_{est}$\\ (\wrt GT)\end{tabular}}}        & \multicolumn{1}{c|}{167.2}            & \multicolumn{1}{c|}{144.09}           & 102.73       & \multicolumn{1}{c|}{235.3}            & \multicolumn{1}{c|}{200.1}            & 163.28       & \multicolumn{1}{c|}{85.85}            & \multicolumn{1}{c|}{94.1}             & 43.76        & \multicolumn{1}{c|}{22.94}                & \multicolumn{1}{c|}{12.97}            & \multicolumn{1}{c|}{13.59}            & \multicolumn{1}{c|}{25.69}        \\ \cline{2-15} 
\multicolumn{1}{|c|}{}                                                                                & \multicolumn{1}{c|}{\textbf{\begin{tabular}[c]{@{}c@{}}$\varDelta_{watt}$\\ (\wrt full-lit)\end{tabular}}} & \multicolumn{1}{c|}{387.2}            & \multicolumn{1}{c|}{387.2}            & 580.8        & \multicolumn{1}{c|}{387.2}            & \multicolumn{1}{c|}{387.2}            & 580.8        & \multicolumn{1}{c|}{387.2}            & \multicolumn{1}{c|}{387.2}            & 580.8        & \multicolumn{1}{c|}{193.6}                & \multicolumn{1}{c|}{387.2}            & \multicolumn{1}{c|}{387.2}            & \multicolumn{1}{c|}{580.8}        \\ \hline
\textbf{}                                                                                             & \textbf{}                                                                                     &                                       &                                       &              &                                       &                                       &              &                                       &                                       &              &                                           &                                       &                                       &                                   \\ \hline
\multicolumn{1}{|c|}{\multirow{3}{*}{\textbf{\begin{tabular}[c]{@{}c@{}}\\\\Luxmeter\\ 11\end{tabular}}}} & \multicolumn{1}{c|}{\textbf{\begin{tabular}[c]{@{}c@{}}$\varDelta_{lux}$\\ (\wrt full-lit)\end{tabular}}}  & \multicolumn{1}{c|}{97.68}            & \multicolumn{1}{c|}{125.15}           & 169.72       & \multicolumn{1}{c|}{167.4}            & \multicolumn{1}{c|}{86.34}            & 194.37       & \multicolumn{1}{c|}{62.67}            & \multicolumn{1}{c|}{118.21}           & 153.02       & \multicolumn{1}{c|}{99.17}                & \multicolumn{1}{c|}{154.28}           & \multicolumn{1}{c|}{167.93}           & \multicolumn{1}{c|}{194.85}       \\ \cline{2-15} 
\multicolumn{1}{|c|}{}                                                                                & \multicolumn{1}{c|}{\textbf{\begin{tabular}[c]{@{}c@{}}$\varepsilon_{est}$\\ (\wrt GT)\end{tabular}}}        & \multicolumn{1}{c|}{194.63}           & \multicolumn{1}{c|}{171.74}           & 131.55       & \multicolumn{1}{c|}{91.14}            & \multicolumn{1}{c|}{128.7}            & 70.21        & \multicolumn{1}{c|}{15.26}            & \multicolumn{1}{c|}{67.87}            & 5.39         & \multicolumn{1}{c|}{9.4}                  & \multicolumn{1}{c|}{241.12}           & \multicolumn{1}{c|}{2.81}             & \multicolumn{1}{c|}{203.69}       \\ \cline{2-15} 
\multicolumn{1}{|c|}{}                                                                                & \multicolumn{1}{c|}{\textbf{\begin{tabular}[c]{@{}c@{}}$\varDelta_{watt}$\\ (\wrt full-lit)\end{tabular}}} & \multicolumn{1}{c|}{387.2}            & \multicolumn{1}{c|}{387.2}            & 580.8        & \multicolumn{1}{c|}{387.2}            & \multicolumn{1}{c|}{387.2}            & 580.8        & \multicolumn{1}{c|}{387.2}            & \multicolumn{1}{c|}{387.2}            & 580.8        & \multicolumn{1}{c|}{193.6}                & \multicolumn{1}{c|}{387.2}            & \multicolumn{1}{c|}{387.2}            & \multicolumn{1}{c|}{580.8}        \\ \hline
\end{tabular}}
\end{table*}

\subsection{The invisible light switch} \label{sec:applications}

The idea behind the Invisible Light Switch is straightforward: the proposed system controls and sets the illumination of the environment by taking into account the information regarding the part of the scene that the user can see or cannot see, by switching off or dimming down the lights outside the user's VFOA, and thus ensuring a consistent energy saving and productivity. 

\begin{figure}[!ht]
	\begin{center}
	    \includegraphics[width=1\linewidth]{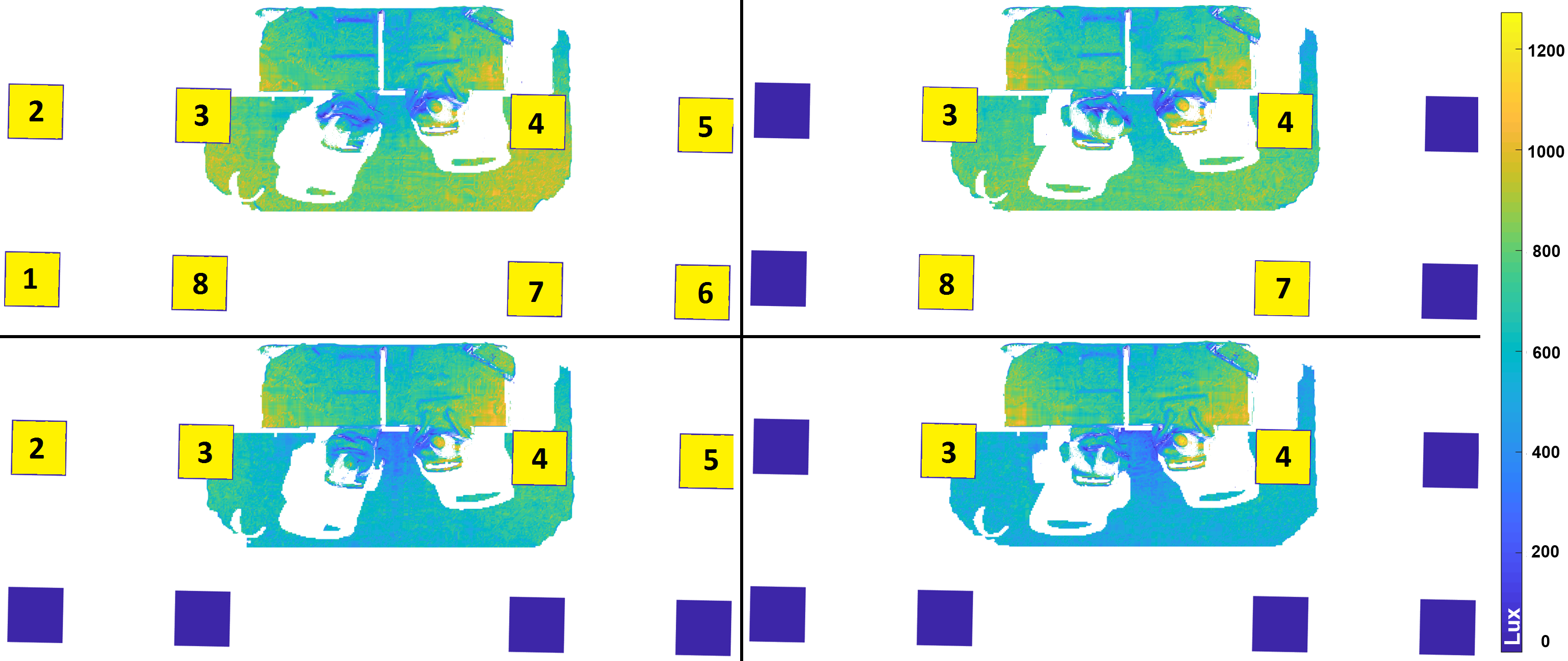}
	\end{center}
	\caption{Qualitative illustration of the VFOA 1 ablation study for Scene 1 presented in table \ref{table:power_saving}. The top left corner shows the illumination map of the full lit case, in comparison to three other light scenarios. As it can be seen the estimated illumination over the desk area where the two occupants have their attention is less affected in comparison to the areas behind them. This show in practice how the invisible light switch application could be established.} \label{fig:test_power_saving}
	\vspace{-8pt}
\end{figure}

In Table \ref{table:power_saving} we examine the applicability of the invisible light switch from the human perspective aspect (luxmeters 10-11) for different head orientation cases (VFOA) in the two scenes. The value $\varDelta_{lux}$ provides the information regarding what is the impact to the light perceived from the occupants (based on the ground truth sensor measurements) on different light source combination scenarios. As we can see this gives us a range of 0-200 lux negative variation even to the most aggressive scenario of having only two luminaires active (the ones to the direct view of the occupants each time). If we connect this with the amount of watts that we can save for this corresponding lighting scenario, \ie $\varDelta_{watt} = 580.8$ watt \wrt to the full lit case, this can give us a total power efficiency of 12379.2 KWatt through a whole day. The value $\varepsilon_{est}$ reports the light estimation error based on our framework, which as we can see again it settles within a range of 0-200 lux overall negative variation. This error shows us how our system aligns with the ground truth measurements, \ie a lower $\varepsilon_{est}$ error the better, and whether the same pattern described above could be followed. A visual example of the VFOA 1 case for scene 1 (see Table \ref{table:power_saving}) can be seen in Figure \ref{fig:test_power_saving}. As it can be easily noticed the estimated illumination over the desk areas have the less affect as we switch off the peripheral light sources and still providing an optimally lit scenario while it is minimally lit.

\section{Conclusion}\label{sec:conclusion}
This paper highlights the importance of a human-centric aided lighting management system which targets productivity over a power saving framework. As a result, in this work we proposed and evaluated a practical (application-wise) system which tries to encapsulate all these three aspects, \ie ambient illumination, human activity and power efficiency. We also for the first time presented a complete system that estimates both the spatial and the individual gaze-gathered light intensity based on a camera-aided solution. We illustrated a possible 66\% of power saving by deploying our framework as the ``Invisible Light Switch'' application which can be used to exploit an optimal illumination pattern for a given human activity.
\newline
\newline
{\noindent\textbf{Acknowledgments:} This project has received funding from the European Union's Horizon 2020 research and innovation programme under the Marie Sklodowska-Curie Grant Agreement No. 676455.}

{\small
\bibliographystyle{ieee}
\bibliography{egbib}
}

\end{document}